\documentclass[sn-aps]{sn-jnl}

\usepackage{bm}
\usepackage{amsfonts,amssymb,pb-diagram}
\usepackage{breqn}

\jyear{2023}%

\theoremstyle{thmstyleone}%
\theoremstyle{thmstyletwo}%

\theoremstyle{thmstylethree}%

\DeclareMathOperator{\sign}{sign}

\newcommand{\argmin}{\operatorname{argmin}}

\newcommand{\by}{\mathbf{y}}

\newcommand{\bbeta}{\bm{\beta}}
\newcommand{\bw}{\bm{w}}

\begin{document}

\title[Neural lasso]{Neural lasso: a unifying approach of lasso and  neural networks}


\author*[1]{\fnm{David} \sur{Delgado}}\email{david.delgado@uc3m.es}

\author[1]{\fnm{Ernesto} \sur{Curbelo}}\email{ecurbelo@est-econ.uc3m.es}

\author[1]{\fnm{Danae} \sur{Carreras}}\email{dcarrera@est-econ.uc3m.es}

\affil*[1]{\orgdiv{Departamento de Estadística}, \orgname{Universidad Carlos III de Madrid}, \orgaddress{\street{Street}, \city{Leganés}, \postcode{28911}, \state{Madrid}, \country{España}}}




\abstract{In recent years, there is a growing interest in combining techniques attributed to the areas of Statistics and Machine Learning in order to obtain the benefits of both approaches. In this article, the  statistical technique lasso for variable selection is represented through a neural network. It is observed that, although both the statistical approach and its neural version have the same objective function, they differ due to their optimization. In particular, the neural version is usually optimized in one-step using a single validation set, while the statistical counterpart uses a two-step optimization based on cross-validation. The more elaborated optimization of the statistical method results in more accurate parameter estimation, especially when the training set is small. For this reason, a modification of the standard approach for training neural networks, that mimics the statistical framework, is proposed. During the development of the above modification, a new optimization algorithm for identifying the significant variables emerged. Experimental results, using synthetic and real data sets, show that this new optimization algorithm achieves better performance than any of the three previous optimization approaches. 
}




\maketitle

\section{Introduction}\label{sec1}

\quad\; Nowadays, there is a growing interest in combining techniques attributed to the areas of Statistics and Machine Learning in order to obtain the benefits of both approaches. 

An example of the above can be found in the area of statistical item response theory, and specifically in the development of computerized adaptive tests \cite{van2013handbook, linden2000computerized}. Yan, Lewis, and Stocking and, later, Ueno and Songmuang proposed the use of decision trees as an alternative to the computerized adaptive tests~\cite{yan2004adaptive, ueno2010computerized}. Later, Delgado-Gomez et al. established mathematically an equivalence between these two techniques that allows the administration of computerized adaptive tests in real-time using item selection criteria that are computationally very intensive \cite{delgado2019computerized}. Recently, several works using neural networks have been published in this field \cite{zhuang2022fully, converse2021estimation}.

Regarding these last works, it is interesting to note the synergies that are being generated between  the areas of Statistics and Neural Networks \cite{cherkassky2012statistics,paliwal2009neural}. Representing statistical models using neural networks provides them with the flexibility and optimization methods of the latter. In a previous pilot study, Laria et al. indicated how the least absolute shrinkage and selection operator (lasso) algorithm can be represented as a neural network \cite{laria2021accurate}. Conversely, linking neural networks to statistical models allows to improve the interpretability of the former~\cite{morala2021towards}. These synergies have occurred in several domains of Statistics such as regression, dimensional reduction, time series, or quality control \cite{de2020statistical}.

In this article, the widely used lasso algorithm is developed from the perspective of neural networks.  To this end, in Section 2, the most relevant features of the lasso algorithm are presented in order to understand the elaboration of its neural version. After that, in Section 3, the entire mathematical formulation proposed by Laria et al. is extended, and the optimization is redefined \cite{laria2021accurate}. Both linear and logistic regressions are considered. In Section 4, several experiments are carried out to evaluate the performance of the neural version and compare it with their statistical counterpart. These experiments are performed on both real and simulated data. Finally, the article concludes in Section 5 with a discussion of the obtained results and future research lines. 

\section{The lasso}

\quad\; Following, the lasso algorithm is briefly presented highlighting the most relevant elements in relation to our proposal. Hereafter, the lasso algorithm will be referred to as \textit{statistical lasso} to differentiate it from its neural version throughout the article. 

\subsection{Formulation}
\quad\; Let $(\bm{x}_i, y_i)$, $i=1, \dots, N$, be a set containing $N$ observations where $\bm{x}_i \in \mathbb{R}^p $  represents the predictors, and $y_i \in \mathbb{R}$ are the associated responses. It is assumed that the predictors are standardized and the responses are centered, i.e., 
\begin{eqnarray}\label{equ:asumpt}
 \sum_{i=1}^N x_{ij} = 0, \hspace{30pt}  \sum_{i=1}^N x_{ij}^2 = 1,  \hspace{30pt}   \sum_{i=1}^N y_i = 0, \hspace{25pt} \text{ for } j=1,2, \dots, p
\end{eqnarray}

The lasso technique was introduced for generalized linear models in the supervised context by Tibshirani \cite{tibshirani2011regression}. It is formulated as the following optimization problem
\begin{eqnarray}\label{equ:lasso}
 \hspace{3pt}\underset{\bbeta}{\argmin}\,
 \mathcal{R}( \bm{y}, \bm{X} \bbeta)
+ \lambda \bigl\| \bbeta \bigr\|_1 
\end{eqnarray}
where $\bm{X}$ is the (standardized) matrix that contains the observations as rows, $\bm{y}$ is the vector
with the corresponding labels, $\bbeta$ is the vector containing the weights of the regression,
and $\lambda \bigl\| \bbeta \bigr\|_1$ is a penalization term. $\mathcal{R} (\bm{y}, \bm{X} \bbeta)$ represents the error term. In this work, we will focus on linear and logistic regression. For linear regression, the error term is given by
\begin{eqnarray}\label{eq:lineal}
    \mathcal{R}_{Lin}(\bm{y}, \bm{X} \bbeta)=\frac{1}{N}\sum_{i=1}^N (y_i-\mathbf{x}_i^t \bbeta)^2
\end{eqnarray}
while the error term for the logistic regression is given by:
\begin{eqnarray}\label{eq:logistic}
\mathcal{R}_{Log}(\bm{y}, \bm{X} \bbeta)= \frac{1}{N} \sum_{i=1}^N\left[ \log(1+e^{\mathbf{x}_i^t \bbeta})- y_i \mathbf{x}_i^t \bbeta \right]
\end{eqnarray}

\subsection{Optimization}

\quad\; Given a fixed $\lambda$, the values of $\bm{\beta}$ are estimated using coordinate descent.  As an example, the coordinate descent update for the $j^{th}$ coefficient in the linear regression case is given by
\begin{eqnarray}
    \hat{\beta}_j= \mathcal{S}_{\lambda}(\frac{1}{N} \langle \mathbf{X}_j, \mathbf{r}_j  \rangle)
\end{eqnarray}
where  $\mathbf{X}_j$ is the $j^{th}$ column of matrix $\mathbf{X}$, the $i^{th}$ component of $\mathbf{r}_j$ is obtained by
\begin{eqnarray}\label{eqn:r}
   \mathbf{r}_j(i)=y_i - \sum_{k \neq j} x_{ik} \hat{\beta}_k  
\end{eqnarray}
and $\mathcal{S}_{\lambda}$ is the soft-thresholding operator defined by
\begin{eqnarray}\label{eqn:soft}
    S_{\lambda}(x)=\sign(x)(|x|-\lambda)_{+}
\end{eqnarray}

The optimal value of $\lambda$  is obtained through a k-fold crossvalidation. A more detailed discussion of the lasso optimization can be found in the book by Hastie, Tibshirani and Wainwright \cite{hastie2015statistical}. A schematic representation of the lasso optimization algorithm is shown in the upper panel of Figure~\ref{fig:laso}.\\

\section{The neural lasso}

\quad\; Similarly to the previous section, the formulation and optimization of the neural lasso is presented.

\subsection{Formulation}

\quad\; Following, the neural representation of the lasso is presented. It begins by presenting the mathematical formulation for linear regression and, afterward, it is extended to logistic regression.\\

\noindent \textbf{Linear regression}\\

\begin{figure}[h!]
		\centering
            \includegraphics[scale=.47]{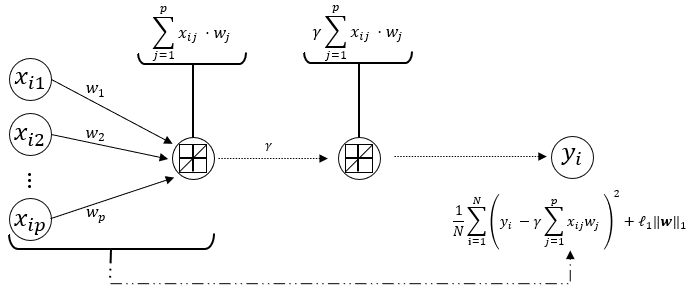}
		\caption{Neural Representation of lasso for linear regression}
		\label{fig:lasso}
\end{figure}

When the error term is given by the mean squared error (MSE), lasso can be characterized as the neural network shown in Figure \ref{fig:lasso}. In this case, the loss function is given by

\vspace{-0.4cm}

\begin{equation}\label{equ:lasso_net_loss}
    \begin{split}
     \mathcal{L}(\bw) &=  
    	\dfrac{1}{N} \sum_{i=1}^{N} \Biggl( y_i -  \gamma \sum_{j=1}^{p} x_{ij} w_j  \Biggr)^2 + \ell_{1} \sum_{j=1}^p \vert w_j \vert \\
           & = \dfrac{1}{N}\| \by - \gamma   \mathbf{X} \bw   \|^{2}_{2} + \ell_{1} \| \bw \|_1
    \end{split}
\end{equation}

where $(\bw,\gamma)$ are the parameters of the network, and $\ell_1$ is a  regularization hyper-parameter. Notice that, by making $\bbeta =\gamma \bw$ and $\lambda=\frac{\ell_{1}}{\gamma}$, equation \eqref{equ:lasso_net_loss} is equivalent to equation \eqref{equ:lasso} using MSE as error term.

An important aspect to keep in mind is that, unlike the statistical lasso, the neural network optimization does not set the weights exactly to zero. Therefore, it is necessary to establish a condition that determines which weights are zeros after each training epoch, and sets them to this value. To do this, we calculate the derivative of the loss function defined in equation \eqref{equ:lasso_net_loss} with respect to $w_j$
\begin{eqnarray}\label{equ:l_derivada1}
\dfrac{\partial \mathcal{L}(\bw)}{\partial w_j}
=\dfrac{-2 \gamma}{N} \sum_{i=1}^{N} \Biggl( y_i -  \gamma  \sum_{k=1}^{p} x_{ik} w_{k} \Biggr) x_{ij} +\ell_1 s_j
\end{eqnarray}
where the term $s_j$ is the subgradient defined by 
\begin{equation}\label{equ:sj}
    s_j =\left\lbrace \begin{array}{cl}
             1 & w_j >0\\
            -1 & w_j <0\\
            \left[-1,1\right] & w_j=0
        \end{array} \right. .
\end{equation}	

\noindent Equation \eqref{equ:l_derivada1} can be rewritten as
\begin{equation}
\dfrac{\partial \mathcal{L}(\bw)}{\partial w_j}=\dfrac{-2 \gamma}{N}
\Biggl(
\sum_{i=1}^{N}  y_i x_{ij}-
 \gamma
\sum_{i=1}^{N} x_{ij}
\sum_{k\neq j} x_{ik}w_k-\gamma  w_j \sum_{i=1}^{N} x_{ij}^2 
  \Biggl)\\+   \ell_1 s_j 
\end{equation}
and, equivalently, in vector form
\begin{eqnarray}
\dfrac{\partial\mathcal{L}(\bw)}{\partial w_j}=\dfrac{-2 \gamma}{N}
\Biggl(
\mathbf{X}_j^t \by-
 \Bigl(\gamma \mathbf{X}_j^t \mathbf{X} \bw^*_j
 -\gamma w_j  \Bigr)+\ell_1 s_j 
\end{eqnarray}
where $\mathbf{X}_j^t$ is the transpose of the $j^{th}$ column of matrix $\mathbf{X}$ (containing observations as rows) and $\bw^*_j$ is the vector $\bw$ with the $j^{th}$ component equal to 0. To obtain the above expression, it has  been taken into account that $\sum_{i=1}^2 x_{ij}^2=1$ since the data are standardized.\\

Equating the derivative to 0 leads to

\begin{eqnarray}
w_j=\dfrac{\dfrac{2}{N}  \gamma \mathbf{X}_j^t \Biggl( \by - \gamma \mathbf{X} \bw^*_j \Biggr) - \ell_1 s_j  }{\dfrac{2}{N} \gamma^2}
\end{eqnarray}

From where it follows that

\begin{equation}\label{eq:condicionL}
\small
    w_j^{op} = \left\lbrace
\begin{array}{cc}
    \frac{\dfrac{2}{N} \gamma \mathbf{X}_j^t \Biggl( \by - \gamma \mathbf{X} \bw^*_j \Biggr) - \ell_1  }{\dfrac{2}{N} \gamma^2} & \textcolor{black}{\text{ if }}  \dfrac{2}{N}\gamma \mathbf{X}_j^t \Biggl( \by - \gamma \mathbf{X} \bw^*_j \Biggr) > \ell_1 \\
    \frac{\dfrac{2}{N} \gamma \mathbf{X}_j^t \Biggl( \by - \gamma \mathbf{X} \bw^*_j \Biggr) + \ell_1  }{\dfrac{2}{N} \gamma^2}, &  \textcolor{black}{\text{ if }} \dfrac{2}{N}\gamma \mathbf{X}_j^t \Biggl( \by - \gamma \mathbf{X} \bw^*_j \Biggr) < -\ell_1 \\
    0 &  \text{ if } \left\vert\dfrac{2}{N} \gamma \mathbf{X}_j^t \Biggl( \by - \gamma \mathbf{X} \bw^*_j \Biggr) \right\vert \leq \ell_1
\end{array}
    \right.
\end{equation}
\\
Indicate that, unlike lasso which needs the three updates of equation \ref{eq:condicionL}, neural lasso only uses the last condition to make weights zero. This is because the update of the weights is performed implicitly during the training of the network. \textcolor{black}{Concisely, after each training epoch, the network will determine if any of the weights can be replaced by 0 by checking if the last condition of the equation \eqref{eq:condicionL} is satisfied using the current estimates}. This difference will be relevant later in the logistic regression.\\

\noindent \textbf{Logistic regression}\\

As shown below, the optimization problem for the logistic case is formulated by
\begin{eqnarray}\label{equ:loss_lasso_log}
\underset{\bbeta}{\argmin} 
\dfrac{1}{N} \sum_{i=1}^N \left[ \log(1+e^{\mathbf{x}_i^t \bbeta + \textcolor{black}{\beta_0}})- y_i \left( \mathbf{x}_i^t \bbeta + \textcolor{black}{\beta_0} \right) \right] + \lambda \bigl\| \bbeta \bigr\|_1
\end{eqnarray}
\\

\begin{figure}[h!]
		\centering
            \includegraphics[scale=.26]{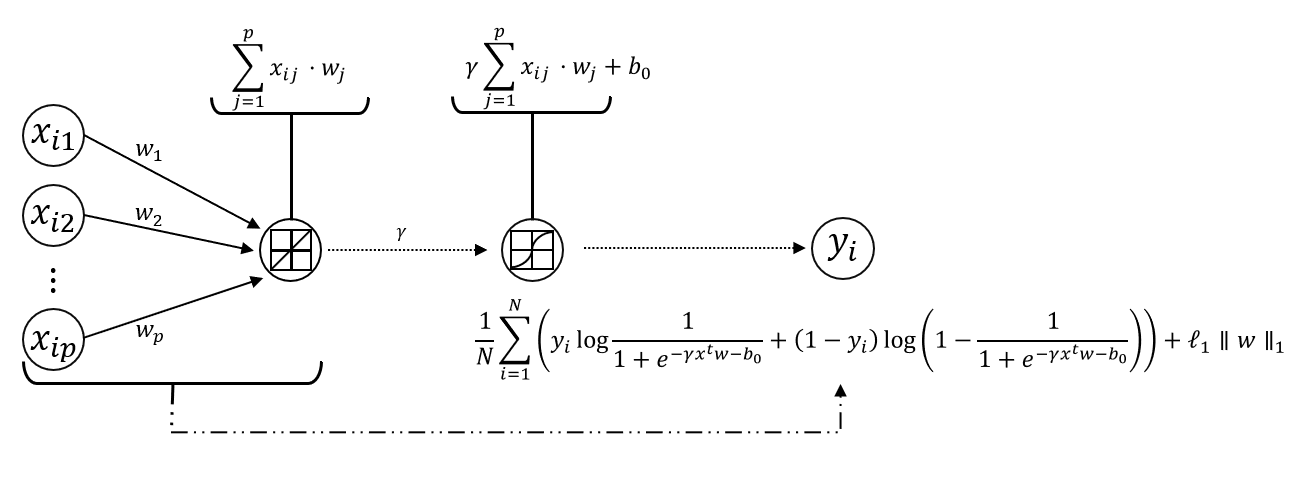}
		\caption{Neural representation of lasso for logistic regression}
		\label{fig:logisticlasso}
\end{figure}

\noindent This problem can be characterized by the neural network shown in Figure \ref{fig:logisticlasso}.

Note that the linear activation of the output layer has been replaced by a sigmoid. In addition, the MSE has been replaced by the binary cross-entropy function whose formula is given by 

\begin{eqnarray}\label{equ:binarycross}
\noindent
-\dfrac{1}{N}\sum_{i=1}^N y_i \log \hat{y}_i  +(1-y_i) \log (1-\hat{y_i})  
\end{eqnarray}

Therefore, the loss function of the network is given by
{
\begin{equation} \label{equ:loss_lassonet_log}
\begin{split}
  \mathcal{L}(\bw)=-\dfrac{1}{N} \sum_{i=1}^N \Biggl( y_i \log \left( \dfrac{1}{1+e^{-\gamma x_i^t \bw - \textcolor{black}{b_0} } } \right)+ (1-y_i) \log\left(1-\dfrac{1}{1+e^{-\gamma x_i^t \bw - \textcolor{black}{b_0} }}\right) \Biggr)\\
  + \ell_{1} \bigl\| \bw \bigr\|_1
\end{split}
\end{equation}

}
It can be seen that equation \eqref{equ:loss_lassonet_log} is equivalent to equation \eqref{equ:loss_lasso_log} as follows. Focusing on the error term of equation \eqref{equ:loss_lassonet_log}:
{\small
\begin{eqnarray*}
\mathcal{R}(\bm{y}, \bm{X} \bw) &= & -\dfrac{1}{N} \displaystyle\sum_{i=1}^N \Biggl( y_i \log \left( \dfrac{1}{1+e^{-\gamma x_i^t \bw - \textcolor{black}{b_0}}} \right)+ (1-y_i) \log\left(\dfrac{1}{1+e^{\gamma x_i^t \bw + \textcolor{black}{b_0}}}\right) \Biggr) \\
&=& -\dfrac{1}{N}  \displaystyle\sum_{i=1}^N \left( -y_i \log(1+ e^{-\gamma\bm{x}_i^t\bw - \textcolor{black}{b_0}}) - (1-y_i)\log(1+e^{\gamma\bm{x}_i^t\bw + \textcolor{black}{b_0}}) \right)\\
&=& \dfrac{1}{N} \sum_{i=1}^{N} \left( y_i \log(1+ e^{-\gamma\bm{x}_i^t\bw - \textcolor{black}{b_0}}) + \log(1+e^{\gamma\bm{x}_i^t\bw + \textcolor{black}{b_0}}) - y_i \log(1+e^{\gamma\bm{x}_i^t\bw + \textcolor{black}{b_0}}) \right)\\
&=&
\dfrac{1}{N}\sum_{i=1}^N \left( y_i  \log \left( \dfrac{1+e^{-\gamma \bm{x}_i^t \bw - \textcolor{black}{b_0}}}{1+e^{\gamma\bm{x}_i^t\bw + \textcolor{black}{b_0}}}\right) + \log(1+e^{\gamma \bm{x}_i^t\bw + \textcolor{black}{b_0}}) \right)\\
&=& 
\dfrac{1}{N}\sum_{i=1}^N \left( y_i \log \left(e^{-\gamma \bm{x}_i^t\bw - \textcolor{black}{b_0}} \right)+ \log(1+e^{\gamma\bm{x}_i^t\bw + \textcolor{black}{b_0}}) \right) \\
&=&\dfrac{1}{N}\sum_{i=1}^N \left( \log(1+e^{\gamma\bm{x}_i^t\bw + \textcolor{black}{b_0}}) - y_i (\gamma\bm{x}_i^t\bw + \textcolor{black}{b_0} ) \right)
\end{eqnarray*}
}
Therefore, \eqref{equ:loss_lassonet_log} becomes
\begin{equation}\label{equ:finalloss_lassonet}
    \mathcal{L}(\bw)=\dfrac{1}{N}\sum_{i=1}^N \left( \log(1+e^{\gamma\bm{x}_i^t\bw + \textcolor{black}{b_0}}) - y_i (\gamma \bm{x}_i^t\bw + \textcolor{black}{b_0}) \right)+ \ell_1 \Vert \bw \Vert_1 
\end{equation}

Defining, as above, $\bbeta= \gamma \bw$, $\lambda =\ell_1/\gamma$ , formulation~\eqref{equ:loss_lassonet_log} is equivalent to formulation \eqref{equ:loss_lasso_log}.\\

Similar to the linear case, it is necessary to establish a mechanism that makes the weights associated with the non-significant variables equal to 0. Taking the derivative of the loss function in equation \eqref{equ:finalloss_lassonet}

\begin{equation}
\dfrac{\partial \mathcal{L}(\bw)}{\partial w_j}
= \dfrac{1}{N}  \sum_{i=1}^N \left( \dfrac{ \gamma x_{ij} e^{\gamma\bm{x}_i^t\bw + \textcolor{black}{b_0}}}{1+e^{\gamma\bm{x}_i^t\bw + \textcolor{black}{b_0}}}-y_i\gamma x_{ij} \right)
+\ell_1 s_j
\end{equation}
\\

Unfortunately, unlike the linear case, it is not possible to isolate the vector $\bw$. The problem is, therefore, approached from a different perspective.\\

Rearranging and equating the above equation to zero

\begin{equation}
  \dfrac{\gamma}{N}  \sum_{i=1}^N \left(\dfrac{   e^{\gamma\bm{x}_i^t\bw + \textcolor{black}{b_0}}}{1+e^{\gamma\bm{x}_i^t\bw + \textcolor{black}{b_0}}}-y_i  \right) x_{ij}
+\ell_1 s_j = 0
\end{equation}
which is equivalent to

\begin{equation}
  \dfrac{\gamma}{\ell_1 N}  \sum_{i=1}^N \left(y_i-\dfrac{   1}{1+e^{-\gamma\bm{x}_i^t\bw + \textcolor{black}{b_0}}}  \right) x_{ij}
 = s_j
\end{equation}

\noindent Following Simon et al. \cite{simon2013sparse}, this is satisfied for $w_j=0$ if

\begin{equation}
  \dfrac{\gamma}{\ell_1 N}  \sum_{i=1}^N \left(y_i-\dfrac{   1}{1+e^{-\gamma\bm{x}_i^t\bw_j^* - \textcolor{black}{b_0}}  }  \right) x_{ij}
 = s_j
\end{equation}
where $\bw_j^*$ is the vector $\bw$ with the $j^{th}$ component equal to 0.
Therefore,

\begin{equation}
  \left\vert \dfrac{\gamma}{\ell_1 N}  \sum_{i=1}^N \left(y_i-\dfrac{1}{1+e^{-\gamma\bm{x}_i^t\bw_j^* - \textcolor{black}{b_0}}}  \right) x_{ij}
 \right\vert= \left\vert s_j\right\vert \leq 1
\end{equation}
Rearranging gives

\begin{equation}
  \left\vert \dfrac{\gamma}{N}  \sum_{i=1}^N \left(y_i-\dfrac{   1}{1+e^{-\gamma\bm{x}_i^t\bw_j^* - \textcolor{black}{b_0}}}  \right) x_{ij}
 \right\vert \leq \ell_1
\end{equation}
which vectorially can be written as

\begin{equation}
 \left\vert \dfrac{ \gamma}{N} \mathbf{X}_j^t \Biggl( \by -  \sigma\left(\gamma \mathbf{X} \bw_j^* + \textcolor{black}{\textbf{b}} \right) \Biggr) \right\vert \leq \ell_1\\
\end{equation}
where $\sigma(x)=1/(1+e^{-x})$ is the sigmoid activation function and $ \textcolor{black}{\textbf{b}}$ is the N-dimensional vector whose all components are equal to $\textcolor{black}{b_0}$.
\\

It is important to note that the way by which neural lasso obtains the  condition that determines whether a weight is zero is different from that of the statistical lasso. The latter uses a quadratic approximation of the error term since it also needs to have an explicit expression of the update of the non-zero weights. Neural lasso only needs to know which weights are zero since the update of the non-zero weights is implicitly performed during the training of the network.

\subsection{Optimization}

\begin{figure}[h!]
		\centering
            \includegraphics[scale=.53]{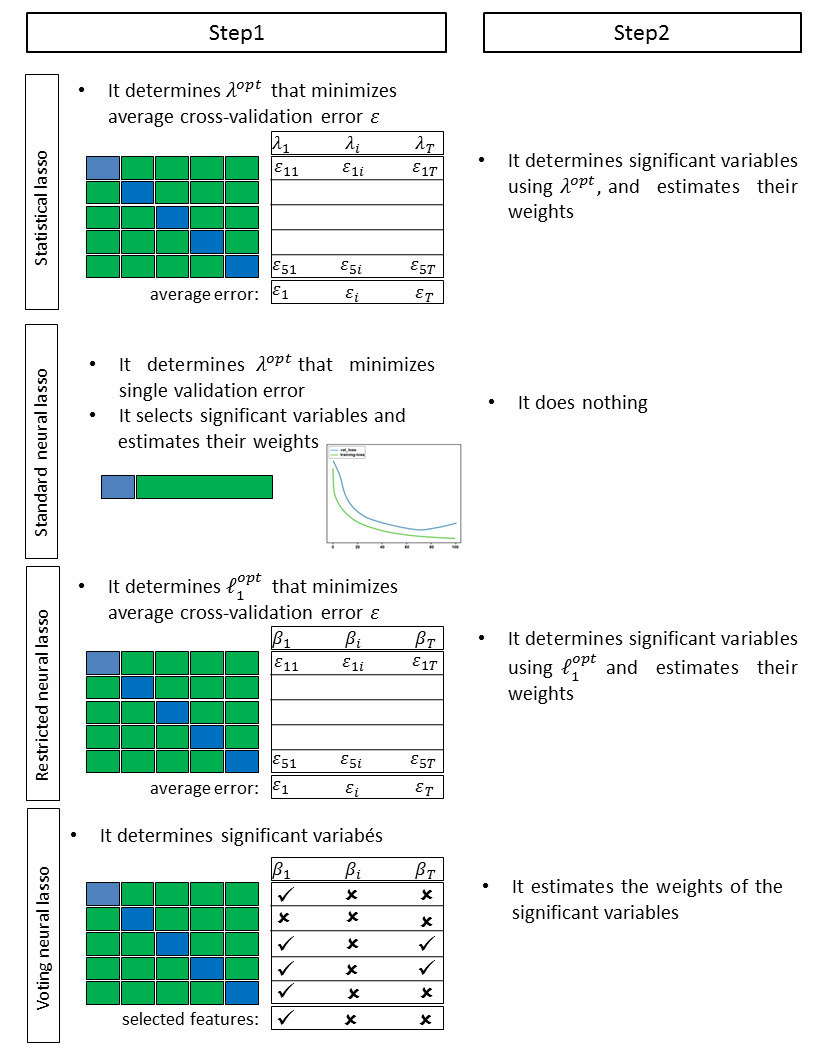}
		\caption{Statiscal lasso and neural lasso algorithms.}
		\label{fig:laso}
\end{figure}

\quad\; An important aspect to discuss is how to estimate the neural lasso weights. In this section, three optimization algorithms are proposed which are shown schematically in the three lower panels of Figure~\ref{fig:laso}.

 Normally, when working with neural networks, its layout is determined by cross-validation and the estimation of its weights by simple validation. That is, once the network layout has been determined, the available data are divided into a training set and a validation set. The training set is used to estimate the network parameters, while the validation set is used to evaluate the performance of the network in an independent set. The resulting network is the one whose weights minimize the validation error. As the network layout is predefined in neural lasso, it is only necessary to estimate its weights using simple validation. This way of training the network will be called \textit{standard neural lasso}.\\

 \textcolor{black}{However, the standard neural lasso may present a disadvantage with respect to the statistical lasso because of how they estimate the weights. The fact that statistical lasso employs cross-validation allows it to use all available observations to obtain an estimate of the error, whereas the standard neural lasso obtains this estimate using only a subset of the observations because it relies on simple validation. For this reason, a second algorithm called \textit{restricted neural lasso} has been developed to train the neural network by mimicking statistical lasso. Restricted neural lasso sets the value of $\gamma$ equal to 1 and establishes it as a non-trainable parameter. Once the $\gamma$ value has been fixed, it also sets the value of the hyper-parameter $\ell_1$ to one of the $\lambda$ values that the statistical lasso considers during its optimization. Having fixed the value of these two parameters, it is possible to perform the cross-validation and the algorithm selects the value of $\ell_1$ that minimizes the cross-validation error. In a second step, the algorithm estimates the weights using the optimal value of $\ell_1$ and setting $\gamma$ equal to 1. Assuming that the network layout is correct, the performance of this second optimization method should be practically identical to that obtained by the statistical lasso.}\\

Finally, during the development of this work, a third optimization approach emerged. This new optimization algorithm, called \textit{voting neural lasso}, combines all the optimization approaches discussed above. Specifically, it uses the cross-validation design used by the restricted neural lasso and by the statistical lasso. However, it does not search for the value of the hyper-parameter $\lambda$ that minimizes the average validation error in the K configurations. For each of the K settings, it selects the value of $\lambda$  with which the smallest validation error is obtained in a similar way to the standard neural lasso. A variable is considered to be significant when it has been selected in most of the K settings. In a second phase, the weights of only these significant variables are estimated without taking into account the penalty term. It is important to note that this approach is not a relaxed lasso \cite{meinshausen2007relaxed}. \\

To summarize the above, three optimization algorithms with three different purposes will be considered. Standard neural lasso obtains the estimation of the weights using the usual procedure of training neural networks.  Restricted neural lasso mimics the statistical lasso method. If these two methods obtain very similar results, a bridge between Statistics and Machine Learning would be built. Finally, voting neural lasso proposes a new way of estimating weights that can be used for both the statistical and the neural versions.\\ 

For the standard neural lasso and for the voting neural lasso, the network is initialized with  $\gamma=1$ and $\ell_1 = \max_j \left\vert \frac{2}{N} \mathbf{X}_j^t  \by \right\vert$ for the linear case and $\ell_1= \max_j \left\vert \frac{1}{N} \mathbf{X}_j^t ( \by -  \sigma(0)) \right\vert $ for the logistic case. In addition, in this article, the Adam optimization algorithm is used to adjust the weights  \cite{kingma2014adam}.\\

\section{Experimental Results}

\quad\; In order to evaluate the performance of the proposed method, three experiments were conducted. The first two focused on the linear case. Specifically, the first one is performed with simulated data and the second one uses several real data sets. The two previous experiments are complemented with a third one aiming to evaluate the proposed method in the logistic case using real data.

\subsection{Experiment 1: Linear case, Simulated data}

\quad\; In the first study, the data were simulated according to the model $ y= \mathbf{X} \bm{\beta} + \epsilon$ where $\mathbf{X}$ is the matrix containing the observations as row, $\epsilon_i \sim N(0,1)$ and
$$ \beta=[1\,2\,3\,4\,\underbrace{0\, \ldots \, 0}_{p-4}]$$
Moreover, the data were simulated from a centered normal distribution so that $\rho_{ij}=0.5^{|i-j|}$ for $1 \leq i <j  \leq p$. In addition, the columns with the predictors were randomly rearranged to avoid possible positional effects.\\

In order to test the performance of the different algorithms, training sets for $p \in \{20,100,200\}$ with sample size $N$ equal to $50$ were generated. For each of the three scenarios, a repeated validation was performed with $100$ runs. In all the repetitions, a test set of $1000$ observations was generated. As performance measures, we calculated the MSE on the test set, the precision (percentage of non-significant variables correctly identified), and the recall (percentage of significant variables correctly identified).
The number of folders K was set to five for the statistical lasso, restricted neural lasso, and voting neural lasso algorithms. Standard neural lasso used $20\%$ of the training data as validation set. Indicate that the analyses using the non-neural versions were performed using the glmnet R package \cite{hastie2021introduction}, while the neural versions were implemented in Pytorch \cite{stevens2020deep}. The obtained results are shown in Table \ref{T50Lineal}. 

\begin{table}[!ht]
\centering
\caption{Results obtained for the linear scenario with synthetic data. For each of the three statistics, the mean and average standard deviation (in parentheses) are shown.  Differences with respect to the statistical lasso algorithm at the 0.05 and 0.01 significance levels are denoted by * and **, respectively.}
\begin{tabular}{clccc}

\hline \hline
      & Method & MSE & Precision & Recall    
\\ \hline \hline
\multirow{4}{*}{p=20}       
    & Statistical lasso  &  1.294 (0.188)  & 0.671 (0.207)  & 1 (0) \\
    & Standard neural lasso & 1.465$^{**}$ (0.341) & 0.644 (0.249) & 1 (0) \\ 
    & Restricted neural lasso  & 1.298 (0.188) & 0.668 (0.210) & 1 (0) \\
    & Voting neural lasso & 1.188$^{**}$ (0.144) & 0.934$^{**}$ (0.072) & 1 (0) \\
\hline 
\multirow{4}{*}{p=100}      
      & Statistical lasso & 1.680 (0.419) & 0.848 (0.087) & 0.998 (0.025)   \\
      & Standard neural lasso & 2.129$^{**}$ (0.789) & 0.808$^{**}$ (0.136)  & 0.998 (0.025) \\ 
      & Restricted neural lasso & 1.695 (0.447) & 0.853 (0.096) & 0.998 (0.025) \\
      & Voting neural lasso  & 1.419$^{**}$ (0.360) & 0.976$^{**}$ (0.017) & 0.998 (0.025) \\
 \hline 
\multirow{4}{*}{p=200}     
    & Statistical lasso  & 1.806 (0.383) & 0.910 (0.053) & 1 (0)
\\
      & Standard neural lasso & 2.338$^{**}$ (0.717) & 0.827$^{**}$ (0.166) & 0.995 (0.035) \\ 
      & Restricted neural lasso & 1.821 (0.395) & 0.910 (0.065) & 1 (0) \\
      & Voting neural lasso  & 1.403$^{**}$ (0.425) & 0.992$^{**}$ (0.007) & 0.990 (0.049) \\
\hline
\end{tabular}\label{T50Lineal}
\end{table}

This table shows that the standard neural lasso performs significantly worse than the non-neural version. As noted above, this is because the standard neural lasso only obtains knowledge of its performance during training on the small validation subset. It is also observed that the performance of the statistical lasso and the restricted neural lasso is almost identical. This confirms that the network design is correct. Finally, a result of interest is that the best results were obtained by the voting neural lasso algorithm which significantly improves those obtained by the three previous approaches. 

\subsection{Experiment 2: Linear case, Real data}

\quad\; The evaluation of the proposed technique was further evaluated using five different real data sets. Specifically, three datasets were obtained from the University of Caroline-Irvine (UCI) repository, and two own datasets were used. The datasets used are the following:

\begin{itemize}
\item[$\circ$] UCI White wine quality  \cite{WineDataset}. This database, containing $4898$ observations, was built to predict the quality of Portuguese ``Vinho Verde" from $11$ predictors. In each of the repetitions, the training set consisted of $4000$ training observations, and the test set was made up of $898$ observations.\\

\item[$\circ$] UCI Boston housing  \cite{harrison1978hedonic}. This dataset consists of $506$ observations with $12$ attributes each. These attributes correspond to the dependent variable, which indicates the median value of owner-occupied homes, and the $11$ predictors used to estimate it. In each of the repetitions, the training set consisted of $400$ training observations, and the test set was made up of $106$.\\

\item[$\circ$] UCI Abalone \cite{nash1994population}. This dataset was collected to predict the age of the abalone from physical measurements. It contains $4177$ observations with nine attributes each. In each of the repetitions, the training set consisted of $3342$ training observations, and the test set was made up of $1935$.\\

\item[$\circ$]  Suicide attempt severity. This database contains information on the severity of $349$ suicide attempts as measured by the Beck suicide intent scale \cite{stefansson2012suicide}. The predictors are $30$ items of the Barrat impulsivity scale \cite{stanford2009fifty}. In each repetition, the training set consisted of $200$ training observations, and the test set was made up of $149$.\\

\item[$\circ$]  Attention Deficit Hyperactivity Disorder (ADHD). It contains the responses provided by $59$ mothers of children with ADHD to the Behavior Rating Inventory of Executive Function-2 containing $63$ items \cite{gioia2015brief}. This dataset has two possible dependent variables measuring the degree of inattention and the degree of hyperactivity of the children as measured by the ADHD rating scale \cite{dupaul1998adhd}.  The training set for each repetition consists of $47$ observations and the validation set consists of $12$ observations.\\

\end{itemize}

As with the previous experiment, $100$ repeated validations are performed, the number of K-folders is set to five, and the validation set contains 20\% of the training data. Obtained results, shown in Table \ref{T50Real}, strengthen the conclusions obtained with synthetic data. In particular, it is observed that the voting neural lasso obtains an MSE similar to that of the statistical lasso but with the advantage of using a significantly smaller number of predictors. It is also observed that the worst performance is obtained with the standard neural lasso. In addition, it can be seen that the statistical lasso and restricted neural lasso obtain practically identical results.

\begin{table}[ht]
\centering
\caption{Results obtained for the linear scenario with real data. For each of the three statistics, the mean and average standard deviation (in parentheses) are shown.  Differences with respect to the statistical lasso algorithm at the 0.05 and 0.01 significance levels are denoted by * and **, respectively.} \label{T50Real}
\begin{tabular}{llcc}
\hline \hline Dataset                                                                        & Method                  & MSE                  & Selected Var. (\%) \\ \hline \hline 
\multirow{4}{*}{\begin{tabular}[c]{@{}l@{}}White wine \\ quality\end{tabular}} & Statistical lasso       & 0.567 (0.027)        & 0.899 (0.087)           \\
                                                                               & Standard neural lasso   & 0.566 (0.027)        & 0.960$^{**}$ (0.073)    \\
                                                                               & Restricted neural lasso & 0.567 (0.027)        & 0.898 (0.084)           \\
                                                                               & Voting neural lasso     & 0.566 (0.028)        & 0.905 (0.070)           \\ \hline
\multirow{4}{*}{\begin{tabular}[c]{@{}l@{}}Boston \\ housing\end{tabular}}     & Statistical lasso       & 25.530 (5.603)       & 0.864 (0.093)           \\
                                                                               & Standard neural lasso   & 25.865 (5.844)       & 0.910$^{**}$ (0.082)    \\
                                                                               & Restricted neural lasso & 25.529 (5.600)       & 0.865 (0.093)           \\
                                                                               & Voting neural lasso     & 25.611 (5.625)       & 0.764$^*$ (0.098)       \\ \hline
\multirow{4}{*}{Abalone}                                                       & Statistical lasso       & 5.063 (0.420)        & 0.981  (0.048)          \\
                                                                               & Standard neural lasso   & 5.334$^{**}$ (0.458) & 0.571$^{**}$ (0)        \\
                                                                               & Restricted neural lasso & 5.061 (0.420)        & 0.981 (0.048)           \\
                                                                               & Voting neural lasso     & 5.060 (0.418)        & 0.964$^*$ (0.062)       \\ \hline
\multirow{4}{*}{\begin{tabular}[c]{@{}l@{}}Suicide \\ attempt\end{tabular}}    & Statistical lasso       & 31.126 (2.380)       & 0.095 (0.123)           \\
                                                                               & Standard neural lasso   & 31.915$^*$ (2.276)   & 0.683$^{**}$ (0.282)    \\
                                                                               & Restricted neural lasso & 31.127 (2.382)       & 0.078 (0.133)           \\
                                                                               & Voting neural lasso     & 31.025 (2.424)       & 0.002$^{**}$ (0.008)    \\ \hline
\multirow{4}{*}{\begin{tabular}[c]{@{}l@{}}ADHD \\ Inattention\end{tabular}}   & Statistical lasso       & 3.616 (1.389)        & 0.257 (0.065)           \\
                                                                               & Standard neural lasso   & 3.680 (1.433)        & 0.334$^{**}$ (0.229)    \\
                                                                               & Restricted neural lasso & 3.614 (1.388)        & 0.252 (0.064)           \\
                                                                               & Voting neural lasso     & 3.787 (1.230)        & 0.145$^{**}$ (0.034)    \\ \hline
\multirow{4}{*}{\begin{tabular}[c]{@{}l@{}}ADHD \\ Hyperactivity\end{tabular}} & Statistical lasso       & 3.465 (1.251)        & 0.312 (0.153)           \\
                                                                               & Standard neural lasso   & 3.883$^*$ (1.686)    & 0.346 (0.205)           \\
                                                                               & Restricted neural lasso & 3.465 (1.259)        & 0.315 (0.159)           \\
                                                                               & Voting neural lasso     & 3.637 (1.198)        & 0.093$^{**}$ (0.029)    \\ \hline 
\end{tabular}
\end{table}


\subsection{Experiment 3: Logistic case, Real data}

\quad\; This last experiment is intended to test the performance of the neural lasso in the logistic scenario. For this purpose, three databases obtained from the UCI repository and one own database are used. A brief description of these databases is given below.\\

\begin{itemize}
\item[$\circ$] UCI Wisconsin Breast cancer \cite{street1993nuclear}. This dataset is composed of $569$ observations. Each observation has $30$ predictors and a dependent variable indicating whether the predictors were obtained from a malignant tumor. The training set was made up of 445 observations while the test set consisted of 124. 
\\

\item[$\circ$] UCI Spam \cite{misc_spambase_94}.
This dataset is made up of $4601$ instances. Each of them contains $57$ predictors and one dependent variable indicating whether the email was spam. The training set consisted of 3975 observations while the test set comprised 626.\\

\item[$\circ$] UCI Ionosphere \cite{sigillito1989classification}. This database is composed of $351$ instances with $34$ predictors and a dependent variable indicating whether the radar signal passed through the ionosphere or not. The training set was made up of 299 observations while the test set consisted of 52.\\

\item[$\circ$] Suicidal Behaviour \cite{blasco2012combining}. This database consists of $700$ observations. Each contains $106$ predictors consisting of responses to items of various scales, and a dependent variable indicating whether the respondent had recently made an attempt.\\
\end{itemize}

The set-up used was similar to that of the two previous sections (K equal to five, $100$ repetitions, and the validation set composed of $20\%$ of the training data). The results obtained are shown in Table~\ref{Logistica}. 

\begin{table}[!ht]
\centering
\caption{Results obtained for the logistic scenario with real data. For each of the two statistics, the mean and average standard deviation (in parentheses) are shown.  Differences with respect to the statistical lasso algorithm at the 0.05 and 0.01 significance levels are denoted by * and **, respectively.} \label{Logistica}
\begin{tabular}{llcc} \hline\hline
Dataset                     & Method                  & ACC                  & Selected Var. (\%)   \\ \hline \hline
\multirow{4}{*}{Cancer}     & Statistical lasso       & 0.963 (0.016)        & 0.359 (0.092)        \\
                            & Standard neural lasso   & 0.964 (0.018)        & 0.160$^{**}$ (0.039) \\
                            & Restricted neural lasso & 0.964 (0.016)        & 0.360 (0.096)        \\
                            & Voting neural lasso     & 0.969$^{**}$ (0.015) & 0.111$^{**}$ (0.018) \\ \hline
\multirow{4}{*}{Spam}       & Statistical lasso       & 0.923 (0.011)        & 0.926 (0.024)        \\
                            & Standard neural lasso   & 0.904$^{**}$ (0.014) & 0.528$^{**}$ (0.056) \\
                            & Restricted neural lasso & 0.924 (0.011)        & 0.927 (0.024)        \\
                            & Voting neural lasso     & 0.915$^{**}$ (0.010) & 0.462$^{**}$ (0.025) \\ \hline
\multirow{4}{*}{Ionosphere} & Statistical lasso       & 0.828 (0.048)        & 0.448 (0.079)        \\
                            & Standard neural lasso   & 0.823 (0.051)        & 0.388$^{**}$ (0.071) \\
                            & Restricted neural lasso & 0.827 (0.047)        & 0.447 (0.080)        \\
                            & Voting neural lasso     & 0.829 (0.048)        & 0.245$^{**}$ (0.040) \\ \hline
\multirow{4}{*}{Suicide}    & Statistical lasso       & 0.650 (0.030)        & 0.093 (0.057)        \\
                            & Standard neural lasso   & 0.627$^{**}$ (0.048) & 0.166$^{**}$ (0.253) \\
                            & Restricted neural lasso & 0.651 (0.029)        & 0.088 (0.061)        \\
                            & Voting neural lasso     & 0.652 (0.031)        & 0.031$^{**}$ (0.010) \\ \hline 
\end{tabular}
\end{table}

Results obtained for the logistic case are similar to those obtained in the linear scenario and presented in the previous two sections. It is observed that the best results are achieved by the voting neural lasso in three of the four settings. A significantly lower accuracy than the statistical lasso is obtained only in the spam data set. It is also observed that the restricted neural lasso and the statistical lasso obtain equivalent results, which again shows the convergence of the neural technique with the statistical one. A small difference, with respect to the results achieved previously, is that the standard neural lasso gets better results than the statistical neural lasso in two settings (Cancer and Ionosphere).

\section{Conclusions}

\quad\; In this work, the lasso algorithm has been implemented by means of neural networks. 
Specifically, the network layout has been defined and three possible optimization algorithms for estimating its weights have been compared. It has been observed that estimating the weights in the way a neural network is usually trained results in poor performance. It has also been shown that it is possible to mimic the optimization of the statistical lasso algorithm with a neural network obtaining almost identical results. The only difference is that the former uses coordinated descent while the latter uses gradient descent. This result brings the fields of Statistics and Machine Learning closer. Finally, an algorithm using a majority vote has been proposed which takes into account how many of the cross-validation scenarios the variable is considered significant. This third algorithm has shown an exceptionally better performance than the widely used statistical lasso. In particular, it has been shown that voting neural lasso either obtains a lower error or obtains a better variable selection in both the linear and logistic cases. Moreover, these results have been obtained using training sets that present a great diversity. They contain a number of observations ranging from only $47$ to $4000$ and a number of predictors varying from $9$ to $200$.

These results open up new lines of research such as developing neural versions of other shrinkage techniques such as the elastic net \textcolor{black}{or extending these algorithms to non-linear versions using the flexibility of neural networks}. It is also important to note that the development of the voting neural lasso has been limited to simple cross-validation which is the information available to the other techniques. However, the use of repeated repetitions or repeated cross-validations, and obtaining confidence intervals, on them might result in a more robust algorithm.   

\section*{Funding}
\quad\; This research was partially funded by: Ministerio de Ciencia e Innovación, Proyectos de Transición Ecológica y Transición Digital  TED2021-130980B-I00, and Instituto Salud Carlos III, grant number DTS21/00091. 

\section*{Data availability}
\quad\;The real data used in this study for the linear regression problem can be obtained from the UCI repository (https://archive.ics.uci.edu/datasets). The real data used for the logistic regression experiment are available from the corresponding author upon request.

\section*{Declarations}
\textbf{Conflict of interest.} The authors have no relevant financial or non-financial interests to disclose.

\bibliography{sn-bibliography}


\end{document}